%% file: acl2024.tex
\documentclass[11pt,a4paper]{article}
\usepackage{acl}
\usepackage{times}
\usepackage{latexsym}
\usepackage{amsthm}

\usepackage{amsmath}
\usepackage{graphicx}
\usepackage{algorithm}
\usepackage[noend]{algpseudocode}

\usepackage{multirow}
\usepackage{adjustbox}
\usepackage{bbm}
\usepackage{wrapfig,lipsum}
\usepackage{xspace}
\usepackage{booktabs}       
\usepackage[normalem]{ulem}
\usepackage{makecell}
\usepackage{amssymb}
\usepackage{pifont}
\usepackage{cleveref}
\usepackage{url}

\title{LLM-A*: Large Language Model Enhanced Incremental Heuristic Search on Path Planning\\
}

\author{
Silin Meng$^{\dagger}$ \ \ \ \ Yiwei Wang$^{\mathsection\dagger}$ \ \ \ \ Cheng-Fu Yang$^{\dagger}$  \ \ \ \
Nanyun Peng$^{\dagger}$ \ \ \ \ Kai-Wei Chang$^{\dagger}$ \\ 
$^\dagger$University of California, Los Angeles \quad $^\mathsection$ University of California, Merced \\
\texttt{silinmeng@cs.ucla.edu}
}

\date{}

\begin{document}
\maketitle
\begin{abstract}
Path planning is a fundamental scientific problem in robotics and autonomous navigation, requiring the derivation of efficient routes from starting to destination points while avoiding obstacles. Traditional algorithms like A* and its variants are capable of ensuring path validity but suffer from significant computational and memory inefficiencies as the state space grows. Conversely, large language models (LLMs) excel in broader environmental analysis through contextual understanding, providing global insights into environments. However, they fall short in detailed spatial and temporal reasoning, often leading to invalid or inefficient routes. In this work, we propose \textbf{LLM-A*}, an new LLM based route planning method that synergistically combines the precise pathfinding capabilities of A* with the global reasoning capability of LLMs. This hybrid approach aims to enhance pathfinding efficiency in terms of time and space complexity while maintaining the integrity of path validity, especially in large-scale scenarios. By integrating the strengths of both methodologies, \textbf{LLM-A*} addresses the computational and memory limitations of conventional algorithms without compromising on the validity required for effective pathfinding. 

\end{abstract}

\input{figures/teaser}
\input{sections/1_introduction}
\input{figures/sample}
\input{sections/2_relatedwork}
\input{figures/algorithm}
\input{sections/3_methodology}
\input{sections/4_experiments}

\input{sections/5_conclusion}

\bibliography{acl2024}
\bibliographystyle{acl_natbib}

\clearpage
\appendix
\input{sections/6_appendix}

\end{document}

%% file: figures/teaser.tex
\begin{figure}[!tb]
	\centering
	\includegraphics[width=1\linewidth] {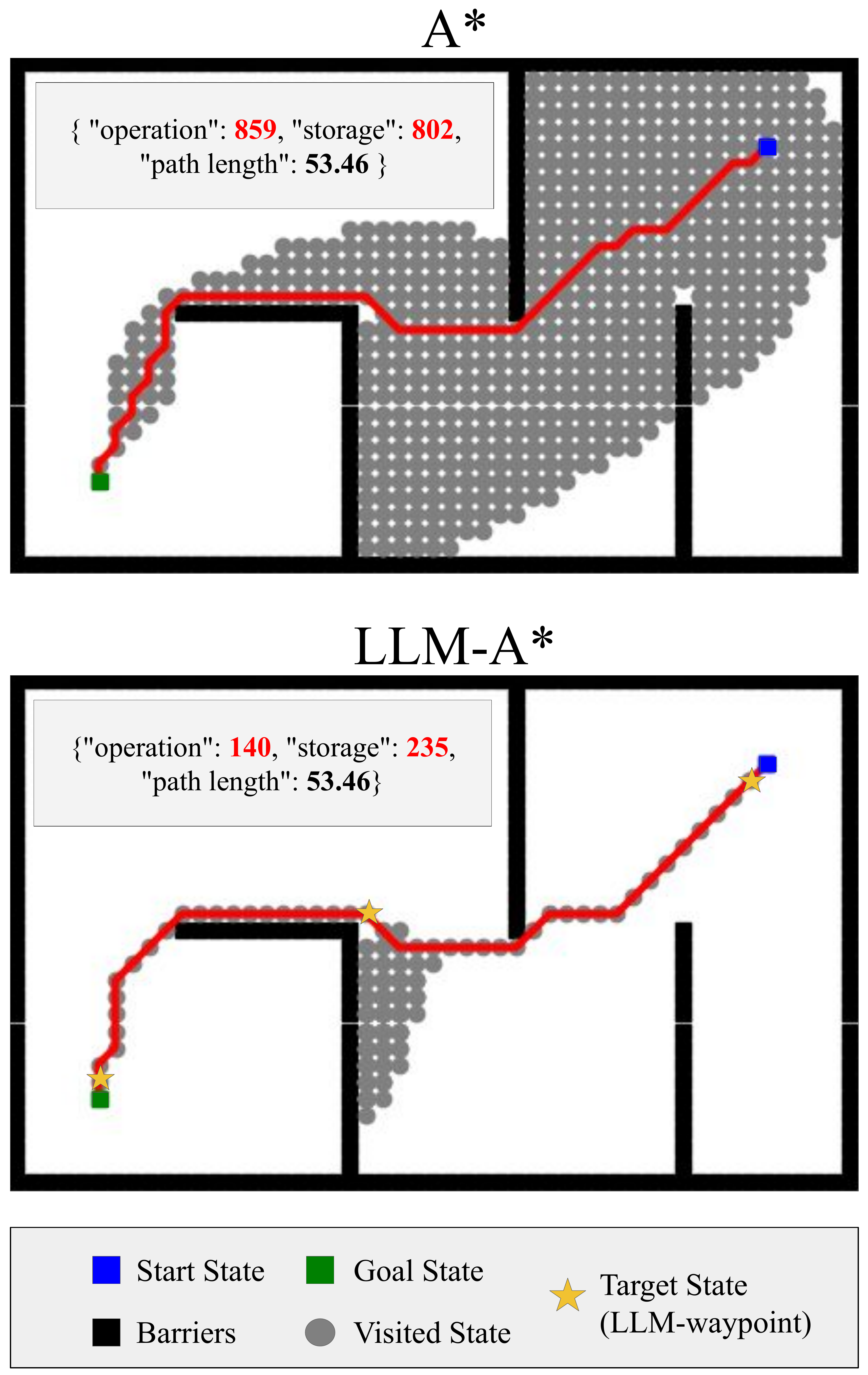}
	\caption{
		An comparison between \textbf{LLM-A*} and A* in computation and memory efficiency during pathfinding process. \textbf{LLM-A*} leverages target states generated by LLMs as waypoints to guide the searching process, significantly reducing the number of visited states, which leads to fewer operations and storage usage than A*.
		\label{fig:compare}}
\end{figure}

%% file: sections/1_introduction.tex
\section{Introduction}

Path planning is the computational process of determining a path from an initial point to a destination point that adheres to specific criteria, such as avoiding obstacles, minimizing travel distance or time, and satisfying other constraints \cite{lavalle2006planning, hart1968formal, karaman2011sampling}. This problem is crucial across several fields, such as robotics, autonomous vehicle navigation, industrial automation, and virtual environment navigation due to its direct impact on the efficiency, safety, and feasibility of operational systems \cite{thrun2005probabilistic, karaman2011sampling, fiorini1998motion, fox1997dynamic}.

Existing path planning algorithms are capable of effectively completing planning tasks and ensuring the validity of their paths. However, as the environment and map scale up, algorithms like A* and its variants \cite{hart1968formal, korf2001time, harabor2011online, jansen2007hpa} encounter an exponential increase in computational and memory demands. This occurs because the pathfinding process can become sub-optimal (see Figure \ref{fig:compare} and \ref{fig:sample}), where the algorithm might spend unnecessary effort exploring less relevant areas, leading to exponential increases in time complexity as the map size enlarges.


Meanwhile, Large Language Models (LLMs) have achieved notable milestones in various planning tasks~\cite{naveed2023comprehensive, yin2023lumos, chen2023fireact, shinn2024reflexion, dou2024reflection}. These models demonstrate capabilities in processing and reasoning over long-context input to provide valuable global insights that reflect their understanding of the environment, such as identifying the relative positions of barriers, agents, and goals. However, they struggle with complex, long-term planning and complex spatial reasoning tasks such as grid-based path planning. LLMs often generate paths that are either invalid or ungrounded, resulting in incomplete or colliding paths, indicating a gap in their capability to handle detailed spatial intricacies \cite{aghzal2023can}.

%% file: figures/sample.tex
\begin{figure*}[!tb]
	\centering
	\includegraphics[width=1\linewidth] {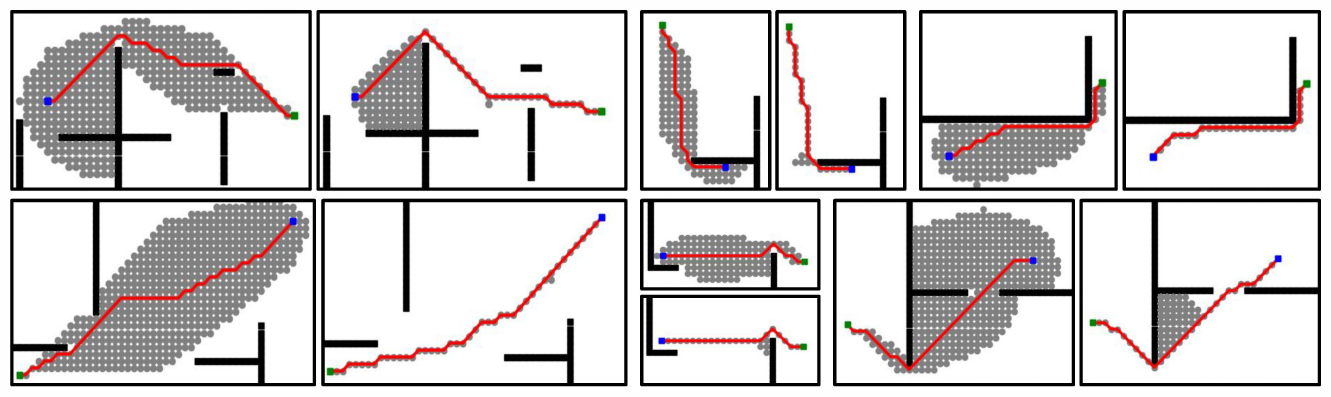}
	\caption{
		Visual comparison of pathfinding efficiency Between A* and \textbf{LLM-A*}. This figure illustrates the performance differences between the traditional A* algorithm (left and upper images) and the \textbf{LLM-A*} algorithm (right and lower images). Red lines indicate the computed paths, blue dots mark the starting state, green dots indicate the goal state, gray areas represent visited states, and black lines denote obstacles. The \textbf{LLM-A*} algorithm demonstrates more efficient pathfinding by requiring significantly fewer visited states than A*.
		\label{fig:sample}}
\end{figure*}

%% file: sections/2_relatedwork.tex
\section{Related Work}

\paragraph{Traditional Algorithms in Path Planning.}
Pathfinding has been pivotal in artificial intelligence, robotics, and computer graphics, with numerous algorithms developed to address various challenges. Among the foundational methods, the A* algorithm, introduced by Hart, Nilsson, and Raphael in 1968, stands out for its use of a heuristic to estimate the cost from the current to the goal node, balancing greedy best-first search with uniform-cost search for efficient pathfinding \cite{hart1968a}. Similarly, Pearl’s Best First Search (BFS), proposed in 1984, prioritizes nodes based on heuristic values but can lead to longer paths due to its focus on the most promising nodes \cite{pearl1984heuristics}.

Extensions of A* have aimed to enhance its efficiency and adaptability. Korf’s Iterative Deepening A* (IDA*), from 1985, combines depth-first search with A*’s heuristic to create a memory-efficient approach \cite{korf1985depth}. Korf also introduced Learning Real-time A* (LRTA*) in 1990, incorporating real-time learning to dynamically update heuristic values, improving performance in changing environments \cite{korf1990real}. Russell’s Simplified Memory Bounded A* (SMA*), from 1992, addresses memory constraints by selectively forgetting less promising paths, making it suitable for resource-limited applications \cite{russell1992memory}.

Further enhancements include Stentz’s Dynamic A* (D*) from 1994, which recalculates paths as the environment changes, proving effective for navigation in unknown or evolving terrains \cite{stentz1994optimal}. Koenig et al.'s Lifelong Planning A* (LPA*), introduced in 2004, incrementally updates paths in dynamic and large-scale environments \cite{koenig2004lifelong}. Harabor and Grastien's Jump Point Search (JPS), proposed in 2011, optimizes A* for only grid-based maps by identifying key "jump points", reducing the number of expanded nodes \cite{harabor2011online}. Nash et al.'s Theta*, from 2007, allows line-of-sight checks between nodes, resulting in more direct paths \cite{nash2007theta}.

Hierarchical approaches, such as Holte et al.'s Hierarchical A* (HA*) from 1996, decompose large pathfinding problems into smaller subproblems through a hierarchy of abstractions, reducing computational complexity \cite{holte1996hierarchical}. Botea et al.'s Hierarchical Path-finding A* (HPA*), introduced in 2004, improves transitions between abstraction levels for efficient large-map pathfinding \cite{botea2004near}.

Specialized methods also contribute significantly. Demyen and Buro's Triangulation-Based Pathfinding (TRA*), proposed in 2006, navigates polygonal environments using triangulation, suited for non-grid-based settings \cite{demyen2006efficient}. Koch's Grid-specific Hierarchical Path-finding (GHPA*), introduced in 2011, optimizes grid maps pathfinding by integrating hierarchical and grid-specific optimizations \cite{koch2011grid}.

\paragraph{Large Language Models in Path Planning.}
Large Language Models (LLMs) have recently achieved remarkable success in natural language processing tasks and other domains \cite{naveed2023comprehensive}. Studies such as \cite{shridhar2020alfworld, song2023llm, shah2023navigation} explore LLMs in high-level planning, highlighting challenges in long-term planning and spatial reasoning \cite{aghzal2023can}. Our research shifts focus to continuous environments, offering a more realistic setting compared to grid-based maps. Continuous spaces align better with real-world conditions, providing a more natural interface for human interaction and allowing higher precision in spatial reasoning.

LLMs show varying proficiency in spatial reasoning \cite{ilharco2020probing, patel2021mapping, bubeck2023sparks, abdou2021can, yang2023planning}, yet face limitations in spatial reasoning and planning \cite{agrawal2023llms, xie2023translating, wu2023reasoning}. We introduce a benchmark for path planning in continuous environments, integrating spatial and temporal reasoning. Prior benchmarks \cite{cote2019textworld, shridhar2020alfred, ruis2020benchmark, wu2021reascan} often neglect temporal planning aspects. Our study further evaluates LLMs in robot motion and path planning contexts, addressing limitations in end-to-end planning \cite{liu2023llm+, chen2023autotamp, xie2023translating, silver2022pddl}.

Recent work, such as the LLM3 framework \cite{wang2024llm} leverages pre-trained LLMs to integrate symbolic task planning with continuous motion generation through motion failure reasoning, where LLM3 iteratively refines both symbolic actions and continuous parameters, significantly improving planning efficiency in dynamic environments, which aligns with our focus on LLMs’ adaptability in correcting low-level planning errors and enhancing resilience in dynamic conditions.

Understanding the interplay between high-level and low-level planning is crucial \cite{latif20243p, yang2023lacma, ding2024mango, zhou2024navgpt}. High-level planning involves strategic goals, while low-level focuses on detailed task execution. Our research explores LLMs' adaptability in correcting low-level planning errors, ensuring resilience in dynamic conditions.

%% file: figures/algorithm.tex
\begin{figure*}[t]
    \centering
    \begin{minipage}{1\linewidth}
        \begin{algorithm}[H]
            \caption{LLM-A* Algorithm for Path Planning}
            \label{alg:llm-a-star}
            \begin{algorithmic}[1]
                \State \textbf{Require:} START state $s_0$, GOAL state $s_g$, OBSTACLE set $obs$, heuristic function $h()$, cost function $g()$, Large Language Model $llm()$

                \State OPEN list $\mathcal{O}_\text{open}\gets\{s_0\}$, CLOSE list $\mathcal{C}_\text{close}\gets\{\}$, TARGET list $\mathcal{T}\gets llm(s_0, s_g, \mathcal{O})$, TARGET state $t\gets\mathcal{T}_\text{start}$, $g(s_0)\gets 0$, $f(s_0)\gets h(s_0)$, $P\gets\{\}$

                \While{$\mathcal{O}_\text{open} \neq \emptyset$}
                    \State $s_a \gets \arg\min_{s \in \mathcal{O}_\text{open}} f(s)$
                    \If{$s_a = s_g$}
                        \State \textbf{return} reconstruct\_path($s_a$)
                    \EndIf
                    \State Remove $s_a$ from $\mathcal{O}_\text{open}$
                    \State Add $s_a$ to $\mathcal{C}_\text{close}$
                    \ForAll{neighbors $s_n$ of $s_a$}
                        \If{$s_n = t$ \textbf{and} $s_g \neq t$}
                            \State $t \gets \mathcal{T}_\text{next}$
                            \State Update $f$-cost of states in $\mathcal{O}_\text{open}$
                        \EndIf
                        \If{$s_n \in \left(\mathcal{C}_\text{close} \cup \text{obs}\right)$}
                            \State \textbf{continue}
                        \EndIf
                        \State Tentative cost $g_\text{tent} \gets g(s_a) + \text{cost}(s_a, s_n)$
                        \If{$s_n \notin \mathcal{O}_\text{open}$ \textbf{or} $g_\text{tent} < g(s_n)$}
                            \State Update path to $s_n$ to go through $s_a$
                            \State $g(s_n) \gets g_\text{tent}$
                            \State $f(s_n) \gets g(s_n) + h(s_n) + \text{cost}(t, s_n)$
                            \If{$s_n \notin \mathcal{O}_\text{open}$}
                                \State Add $s_n$ to $\mathcal{O}_\text{open}$
                            \EndIf
                        \EndIf
                    \EndFor
                \EndWhile
                \State \textbf{return} \textbf{failure}
            \end{algorithmic}
        \end{algorithm}
    \end{minipage}
\end{figure*}

%% file: sections/3_methodology.tex
\section{Methodology}
\subsection{A* Algorithm}

The A* algorithm is a widely used pathfinding and graph traversal algorithm. It seeks to find the shortest path from a start node $s_0$ to a goal node $s_g$ by combining the strengths of Dijkstra's Algorithm and Greedy Best-First Search.

A* employs a heuristic function $h(s)$ to estimate the cost from a node $s$ to the goal, and a cost function $g(s)$ to track the exact cost from the start to $s$. The total cost function $f(s)$, defined as $f(s) = g(s) + h(s)$, guides the search towards the goal. The algorithm operates as follows:

\begin{enumerate}
\item \textbf{Initialization:} Place the start node $s_0$ in the OPEN list with $f(s_0) = g(s_0) + h(s_0)$, and initialize the CLOSED list as empty.
\item \textbf{Search:} Continuously select the node $s$ from the OPEN list with the lowest $f$-cost, expand its neighbors, and update their costs. If a neighbor $s_n$ offers a cheaper path than previously recorded, update its cost and parent node. Repeat until the goal node $s_g$ is reached or the OPEN list is empty.
\item \textbf{Path Reconstruction:} Once $s_g$ is reached, reconstruct the path by tracing back from $s_g$ to $s_0$ via parent nodes.
\end{enumerate}

The heuristic $h(s)$ should be admissible, meaning it does not overestimate the true cost to reach the goal. This ensures the path optimality of A*.

\subsection{LLM-A* Algorithm}
\textbf{LLM-A*} integrates the global insights provided by Large Language Models (LLMs) with the A* algorithm’s optimal local search mechanism, where achieves a balance between the efficiency of the pathfinding process and optimality. The pseudocode for \textbf{LLM-A*} is shown in Algorithm \ref{alg:llm-a-star}, and it closely resembles the original A* algorithm.

\textbf{LLM-A*} accepts the same inputs as A*, with the addition of an obstacle state variable, denoted as $obs$. This obstacle state is utilized to compute a TARGET list $T$, which comprises a sequence of path nodes from the start state $s_0$ to the goal state $s_g$. This list is generated through a prompt to a large language model, reflecting the model's understanding and global perspective of the current environment. The returned $T$ must meet two critical constraints in the following:

\begin{enumerate}
    \item \textbf{Containment of Start and Goal Points:} $T$ must include the start point and goal point that match the inputs $s_0$ and $s_g$. If the returned $T$ does not satisfy this requirement, $s_0$ and $s_g$ must be inserted by algorithm.
    \item \textbf{Obstacle Avoidance:} Every target node $t$ in $T$ must not be located within any obstacle $obs$. If any node $t$ is found within an obstacle, it is removed from $T$ by algorithm.
\end{enumerate}
The pathfinding process of \textbf{LLM-A*} is similar to that of A*. It uses a heuristic function $h$, a cost function $g$, an OPEN list $O$, and a CLOSED list $C$. The algorithm searches through each state in $O$ until the goal state $s_g$ is reached. Each explored state $s_a$ is saved into $C$ to avoid redundant searches. The distinction that encapsulates the main differences between \textbf{LLM-A*} and A* happens during the expansion of the neighbor state $s_n$ (see in Algorithm \ref{alg:llm-a-star}:$13$-$15$). For each $s_n$, we check if it matches the current target $t$ from $T$. If the current $t$ is reached, $t$ is updated to the next target in $T$. Subsequently, the $f$-cost of every state in the current $O$ is re-computed, where the $f$-cost in \textbf{LLM-A*} is computed as the sum of the state's cost, the heuristic value, and the cost from the state to current $t$ (see in Algorithm \ref{alg:llm-a-star}:$20$), defined as $f(s) = g(s) + h(s) + cost(t, s)$. This step introduces an additional computational amount to the pathfinding process, and the time complexity scales linearly with both the length of $T$ and the increasing size of $O$. However, it is important that this re-computation process ensures that the $f$-cost of visited states in $O$ remains accurate and updated with the new target $t$.

\paragraph{General Applicability.} 
\textbf{LLM-A*} retains the versatility of the original A*, making it suitable for a wide range of pathfinding tasks across various environments, where specialized A* variants such as JPS and GHPA* \cite{harabor2011online, koch2011grid}, which are tailored to grid maps and specific scenarios, and the mechanism of \textbf{LLM-A*} is able to handle diverse and large-scale environments effectively. This generality positions \textbf{LLM-A*} as a robust alternative to A*.

\subsection{Prompt Techniques} \label{sec:prompt}

\paragraph{Few shot Learning.} In the methodology we termed "Few Shot Learning" or "Vanilla Prompting," our initial approach involves directly presenting the Large Language Model (LLM) with ground-truth sequences of actions as prompts. This method is informed by previous studies which have demonstrated that the performance of such models can vary significantly based on the volume of task-specific examples provided \cite{cao2019theoretical, razeghi2022impact}. To investigate this further, we employed a few-shot learning technique, wherein we provides five demonstrations (See Table \ref{fig:fewshot} in Appendix) presented to the LLM. This approach aimed to determine the optimal number of examples that would enhance the model's accuracy and learning efficiency.
\paragraph{Chain of Thought.} The Chain-of-Thought (CoT) methodology, as proposed by \cite{wei2022chain}, introduces a technique that encourages a Large Language Model (LLM) to engage in a sequential, step-by-step reasoning process. This approach has demonstrated substantial efficacy in tasks necessitating multiple layers of reasoning and decision-making. In light of its proven effectiveness, we have adapted the CoT strategy (See Table \ref{fig:coT} in Appendix) to the specific requirements of path planning.
\paragraph{Recursive Path Evaluation.} The Recursive Path Evaluation (RePE) methodology (See Table \ref{fig:repe} in Appendix) is designed to guide Large Language Models (LLMs) in generating paths incrementally, with a particular emphasis on evaluating each step in the process. This approach gains its effectiveness from deconstructing the path planning problem into three distinct sub-problems: environment analysis, path generation, and path evaluation. By following these sub-problems in a recursive manner, the model systematically navigates towards the goal, ensuring compliance with predefined constraints at each stage. This concept bears a resemblance to the ReAct approach, Step Back QA, and Self Reflection \cite{yao2022react, zheng2023take, renze2024self} in its processing step by step foundational principles. Meanwhile, RePE receives no feedback or observation from environment, and it distinctively focuses on a step-by-step progression and only intrinsic reasoning, where the path is constructed one point at a time with environment analysis and path evaluation. This methodical approach not only facilitates more precise navigation by the LLM but also allows for continuous assessment and adjustment at each juncture, thereby may enhancing the overall accuracy of the path planning process.

%% file: sections/4_experiments.tex
\section{Experiments}
\subsection{Dataset} \label{sec:dataset}
Our dataset consists of $100$ manually selected $50 \times 30$ maps from a randomly generated collection, each with 10 different start and goal positions. Therefore, there are $1000$ samples in total (see Figure \ref{fig:compare} for sample visualization). Our data conform to the standard of search-based algorithm environments in a continuous space. Each map includes the following parameters: 

\begin{itemize}
    \item \textbf{$x\_range$}: The minimum and maximum x-coordinates of the environment boundary range as $[x\_min, x\_max]$.
    \item \textbf{$y\_range$}: The minimum and maximum y-coordinates of the environment boundary range as $[y\_min, y\_max]$.
    \item \textbf{$horizontal\_barriers$}: List of horizontal barriers, each represented as $[y, x\_start, x\_end]$.
    \item \textbf{$vertical\_barriers$}: List of vertical barriers, each represented as $[x, y\_start, y\_end]$.
    \item \textbf{$start\_goal$}: List of $10$ unique start and goal positions for each map.
\end{itemize}
These parameters define the structure and constraints of each map, ensuring consistency and relevance to the standard experimental environment conditions for search-based algorithms. Notably, the discretization of points and actions within this continuous framework is a necessary simplification that allows us to make the problem tractable and effectively evaluate algorithms. Meanwhile, the map environment is able to scale properly for scalability experiment.

\subsection{Experimental Setup}
\paragraph{Large Language Model.} 
We employ \textbf{GPT-3.5-TURBO} and \textbf{LLAMA3-8B-16bit} for their balance of robustness and cost-effectiveness in validating the \textbf{LLM-A*} algorithm. Prompts include simple instructions, standard 5-shot examples, chain of thought with 3-shot, and recursive path evaluation with 3-shot for in-context learning (see Section \ref{sec:prompt}).

\paragraph{Experiment Environment.} 
Our system facilitates search-based pathfinding within a scalable framework designed for environments of varying complexity. It consists of modules for environment management, agent control, and visualization (see Section \ref{sec:dataset}).
\begin{itemize}
\item \textbf{Environment Management:} Configures the environment and provides feedback, ensuring a challenging setup.
\item \textbf{Agent Control:} Customizes the agent's operations using the algorithm and model, operating on discrete points and actions to make the problem tractable.
\item \textbf{Visualization:} Offers real-time and final visual outputs for comprehensive analysis.
\end{itemize}

While the environments considered are presented within a continuous framework, both the LLM and A* algorithms operate on discrete points and actions. This necessary simplification allows us to effectively evaluate the proposed LLM-A* algorithm, ensuring the system remains applicable to a wide range of complex environments.

\input{tables/experiment}
\paragraph{Experiment Subject.} Our experiments focus on two critical aspects: efficiency and scalability. For \textbf{efficiency}, we assess the number of operations and the storage required for the pathfinding process, defined as time and space complexity, respectively. Additionally, we evaluate the generated path length to assess path efficiency. These metrics are used to compute a composite efficiency score, as presented in Table \ref{tab:experiment}. Larger environments and maps are employed to better illustrate algorithm efficiency, as they offer a more comprehensive reflection of the algorithm's performance under increased complexity. Specifically, we conducted efficiency experiments on a $50 \times 30$ map of the original sample size. This size was selected as it provides a substantial basis for evaluating efficiency while keeping the computational demands within a manageable range. Beyond this scale, the experiment run times become excessively long. For \textbf{scalability}, we tested both A* and LLM-A* algorithms across 10 different scales, from 1 to 10, to examine how they adapt to progressively larger environments, as depicted in Figure \ref{fig:increment}.

\subsection{Evaluation Metrics} \label{sec:metric}
We assess \textbf{LLM-A*} against A* using metrics for operation efficiency, storage efficiency, and path quality. Performance is summarized by the geometric mean of performance ratios between \textbf{LLM-A*} and A* for operation, storage, path length, offering a balanced view less affected by outliers.

\paragraph{Operation and Storage Ratios.} 
We compute the geometric mean of the ratios of operations and storage used by \textbf{LLM-A*} relative to A* (\(\frac{\text{LLM-A*}}{\text{A*}}\)). A lower score indicates better efficiency, e.g., a $50\%$ score means \textbf{LLM-A*} uses $50\%$ of the resources compared to A*.

\paragraph{Relative Path Length.} 
Path quality is evaluated by comparing the path lengths from \textbf{LLM-A*}, A* and LLM-only approach to the optimal path. The geometric mean of these ratios indicates how close \textbf{LLM-A*} paths are to optimal.

\paragraph{Valid Path Ratio.} 
This metric measures the proportion of successful pathfinding attempts, often indicating that the generated path is collision-free and completable. A higher ratio indicates better reliability, showing the algorithm’s effectiveness in generating valid paths consistently.

\paragraph{Growth Factor.} 
We assess how performance scales from a $50 \times 30$ environment to larger sizes by calculating the arithmetic mean of the growth factors for operations and storage. This normalizes efficiency and scalability across different environment sizes.

\subsection{Quantitative Analysis}\label{sec:result}
Table \ref{tab:experiment} presents a comparative analysis of three pathfinding methodologies: the classical A* algorithm, an LLM-only approach, and our proposed \textbf{LLM-A*} approach. The A* algorithm serves as the baseline, with an index value of $100$ indicating performance equivalent to A*, as outlined in Section \ref{sec:metric}. The methodologies are evaluated on maps $50 \times 30$ of original map sizes.

The results demonstrate that \textbf{LLM-A*} significantly enhances both operation and storage efficiencies compared to A*. Specifically, when utilizing the \textbf{LLM-A*} model, \textbf{GPT-3.5} achieves a $57.39\%$ score in operations and a $74.96\%$ score in storage, with a modest $2.44\%$ increase in relative path length. Superior, with the \textbf{LLAMA3} model, \textbf{LLM-A*} reduces operations by $44.59\%$ and storage by $64.02\%$, accompanied by a slight $2.47\%$ increase in relative path length. These results highlight that \textbf{LLM-A*} not only reduces resource consumption but also maintains path validity, consistently achieving a valid path ratio of $100\%$ across all scenarios. The observed increase in path length remains relatively low compared to the optimal path.

When compared to other variants using non-admissible heuristics, LLM-A* demonstrates an superior performance in term of operation and storage efficiency. Dynamic weighted A* (with initial weight of 2 and 0.99 decay) employs a static logic to dynamically update the weight, but it lacks the flexibility inherent to LLM-A*. Consequently, dynamic weighted A* falls short in terms of both operation efficiency and storage efficiency compared to our advanced LLM-A* approach. 

Meanwhile, the LLM-only approach underperforms compared to \textbf{LLM-A*} and A* algorithms in terms of both path efficiency and validity. When used in isolation, LLMs may struggle with comprehensive path planning due to their lack of heuristic guidance, which is provided by \textbf{LLM-A*}, or the deterministic guarantees inherent in A*. The integration of LLM insights in \textbf{LLM-A*} significantly enhances its operational and storage efficiencies, surpassing the performance of A*.

\paragraph{Ablation Analysis.} The Recursive Path Evaluation (RePE) prompting method achieves marginal improvements in relative path length for the \textbf{LLM-A*} approach using \textbf{GPT-3.5}, with an increment of $2.41\%$. This suggests some potential of RePE's step-by-step progression and intrinsic reasoning capabilities in generating more optimal waypoints, resulting in slightly more efficient paths. However, it is important to acknowledge that RePE underperforms compared to Chain of Thought (CoT) and few-shot prompting in a both operation and storage ratio, as well as efficiency in the LLM-only approach. This observation aligns with the limitations of LLMs in executing end-to-end path planning and spatial-temporal reasoning, which can affect their proficiency in sequentially reasoning out detailed path sequences and lead to issues such as hallucinations and misunderstandings, highlighting the diminish of  RePE's efficiency for long-horizon tasks, where the intermediate points chosen using this technique are not optimal. Therefore, while RePE shows some promise, its overall effectiveness is limited compared to other methods in both LLM-A* and LLM-only scenarios.

\input{figures/increment}
\paragraph{Scalability Analysis.} 
Figure \ref{fig:increment} provides a comparative analysis of the computational and memory efficiency of the A* and LLM-A* algorithms across environments of different scales. The analysis is presented through two metrics: the growth factor of operations and the growth factor of storage, with respect to different environment scales.

The results from Fig.~\ref{fig:increment} indicate that \textbf{LLM-A*} significantly outperforms \textbf{A*} in both computational and memory efficiency across various environment scales. While A* grows exponentially in operations and storage, LLM-A* achieves near-linear scalability relative to the environment size. This performance advantage arises from the learning-based enhanced heuristic values incorporated into LLM-A*, which allow it to avoid unnecessary node exploration and facilitate a more direct search towards the goal. This adaptation proves especially effective in larger and more complex environments. The efficiency gains of LLM-A* are particularly noteworthy in environments scaled up to 10 times, where the inefficiencies of A* become increasingly pronounced.

\input{figures/showcase}

\subsection{Qualitative Analysis}\label{sec:qualitative}

From the visualization in Figure \ref{fig:compare}, \textbf{LLM-A*} identifies the optimal path with only 140 operations, less than one-fifth the 859 operations required by A*, as well as the storage reduction. Both algorithms utilize a priority queue that stores the $f$-cost of each reached state, with the state having the lowest $f$-cost selected for exploration. The fundamental distinction between the two algorithms lies in their calculation of the $f$-cost or heuristic values. In addition, as the map size increases, this operational efficiency difference could become more pronounced, as further illustrated in Figure \ref{fig:increment}.


As illustrated in Figure \ref{fig:showcase}, \textbf{LLM-A*} leverages heuristic values derived from LLM-generated waypoints in addition to standard heuristic from A*, resulting in a dynamic heuristic that changes as the algorithm progresses. This dynamic adjustment is achieved through switching to the next target state during search when the current target state is reached. Each time the target state changes, the heuristic values for all previously reached states are recalculated. This allows \textbf{LLM-A*} to steer the search direction towards areas deemed more favorable by the large model at various stages of the search.

In contrast, A* employs a static heuristic for each state, which remains unchanged throughout the search. This static approach can lead to extensive exploration of non-optimal paths, including dead-end areas in the environment.

%% file: tables/experiment.tex
\begin{table*}[ht!]
    \centering
    \begin{adjustbox}{width=1\linewidth}
        \begin{tabular}{@{}lll|c c c | c@{}}
            \toprule
            \textbf{Methodology} 
            & \textbf{Base Model} 
            & \textbf{Prompt Approach}
            & \textbf{Operation Ratio $\downarrow$ (\%)} 
            & \textbf{Storage Ratio $\downarrow$ (\%)} 
            & \textbf{Relative Path Length $\downarrow$ (\%)} 
            & \textbf{Valid Path Ratio $\uparrow$ (\%)} \\
            \midrule
            \multirow{1}{*}{\textbf{A*}} 
            & - & - & 100 & 100 & 100 & 100 \\
            \multirow{1}{*}{\textbf{Dynamic WA* $(w=2)$}} 
            & - & - & 60.91 & 78.53 & 100.24 & 100 \\
            \midrule
            \multirow{6}{*}{\textbf{LLM}} 
            & \multirow{3}{*}{GPT-3.5} 
            & Few-Shot & - & - & 119.38 & 12.80\\
            &  & CoT & - & - & 151.73 & 15.20 \\
            &  & RePE & - & - & 183.87 & 7.80\\
            \cmidrule{2-7}
            & \multirow{3}{*}{LLAMA3} 
            & Few-Shot & - & - & 111.05 & 12.60\\
            &  & CoT & - & - & 114.89 & 12.00 \\
            &  & RePE & - & - &  138.32 &  16.40\\
            \midrule
            \multirow{6}{*}{\textbf{LLM-A*} (Ours)} 
            & \multirow{3}{*}{GPT-3.5} 
            & Few-Shot & 57.39 & 74.96 & 102.44  & \textbf{100}\\
            &  & CoT & 69.50 & 83.65 & 102.54 & \textbf{100}\\
            &  & RePE & 85.47 & 96.53 & \textbf{102.41}  & \textbf{100} \\
            \cmidrule{2-7}
            & \multirow{3}{*}{LLAMA3} 
            & Few-Shot & \textbf{44.59} & \textbf{64.02} & 102.47  & \textbf{100}\\
            &  & CoT & 47.60 & 66.27 & 102.46  & \textbf{100}\\
            &  & RePE & 64.08 & 80.19 & 102.54  & \textbf{100}\\
            \bottomrule
        \end{tabular}
    \end{adjustbox}
    \caption{Quantitative analysis of three pathfinding methodologies: the classical A* algorithm, dynamic weighted A* with initial weight of 2 and 0.99 decay, an LLM-only approach, and our proposed \textbf{LLM-A*} approach. The methodologies are evaluated on the map size ($50 \times 30$) of original samples. The LLM-only approaches explore the path without explicitly searching the space grid by grid, so we do not report the operation and storage ratio. The table includes the results from \textbf{GPT-3.5} and \textbf{LLAMA3} models with three prompting approaches: Few-Shot, Chain of Thought (CoT), and Recursive Path Evaluation (RePE) for both LLM-only and \textbf{LLM-A*} approaches (see Section \ref{sec:result} for details).}
    \label{tab:experiment}
    \vspace{-0.4cm}
\end{table*}

%% file: figures/increment.tex
\begin{figure}[!tb]
	\centering
	\includegraphics[width=1\linewidth] {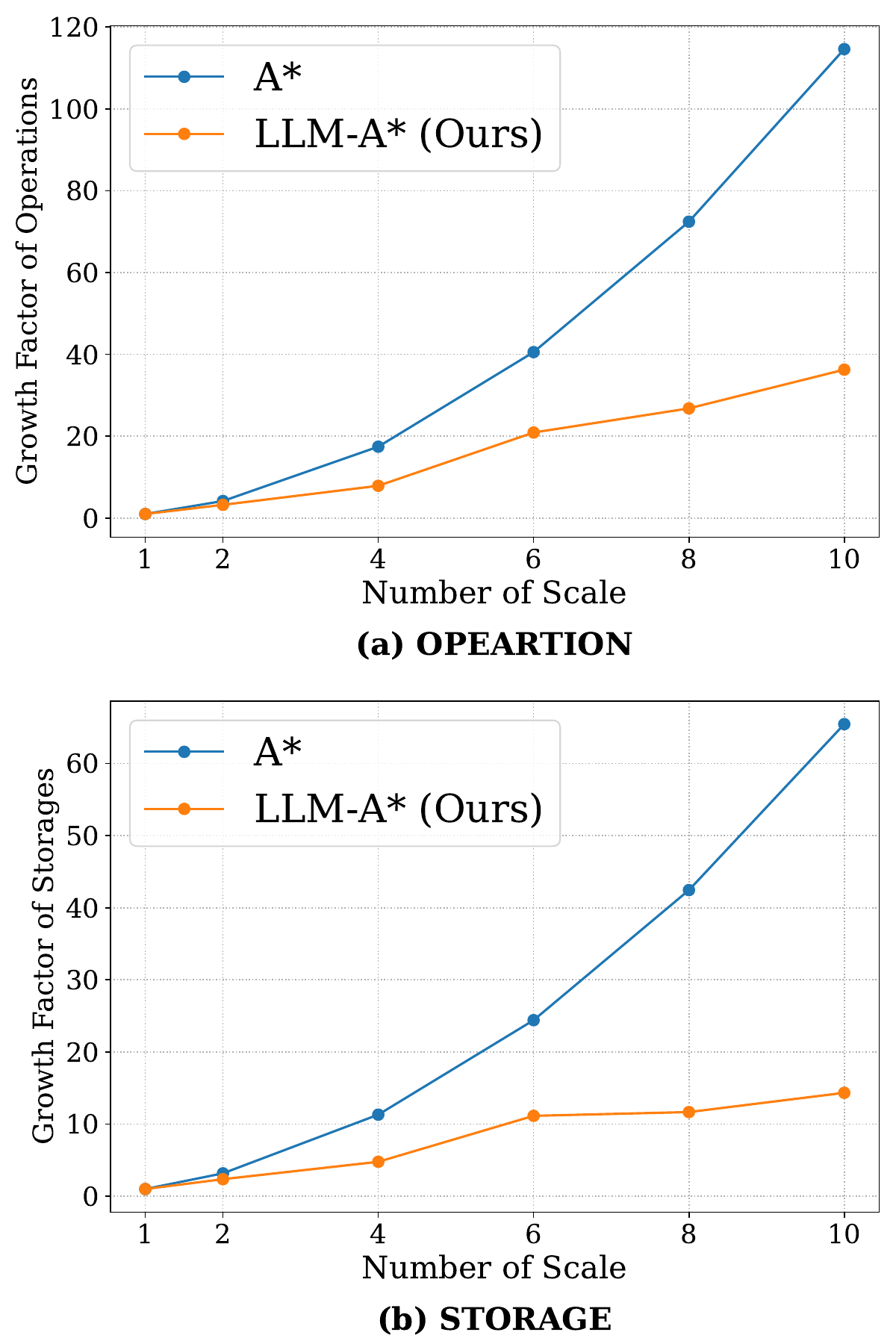}
	\caption{
            The comparative analysis examines the computational and memory efficiency between A* and \textbf{LLM-A*} (incorporating \textbf{LLAMA3} with few-shot prompting) across scaled environments ranging from $1$ to $10$ times enlargement, based on the means of $10$ trials of random sampling. A* exhibits exponential growth in both (a) OPERATION and (b) STORAGE with linear increasing, environment scale, in contrast, \textbf{LLM-A*} achieves a near linear scalability.
		\label{fig:increment}}
\end{figure}

%% file: figures/showcase.tex
\begin{figure}[!tb]
	\centering
	\includegraphics[width=\linewidth] {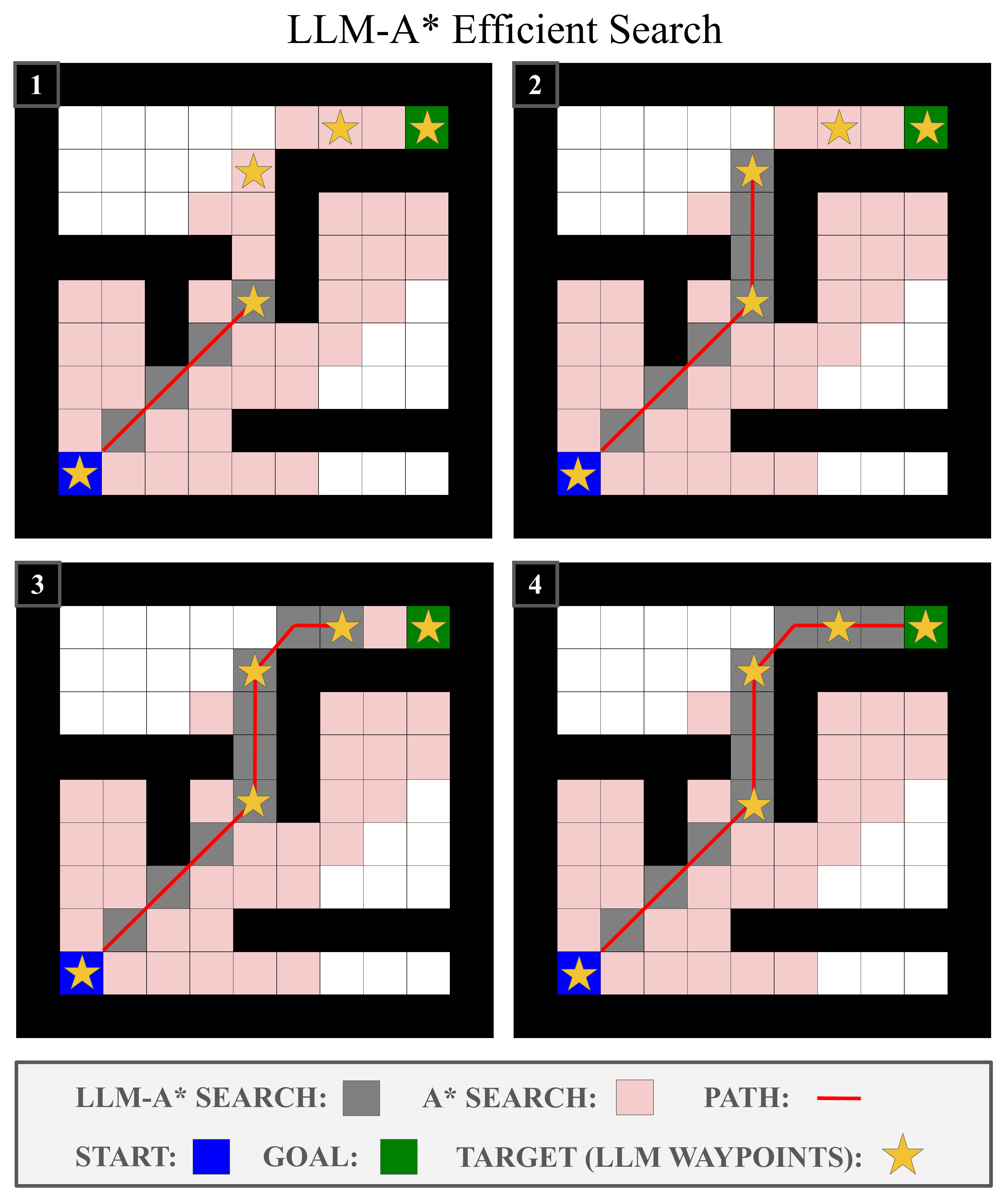}
	\caption{
Visualization of pathfinding process with \textbf{LLM-A*} algorithms (under \textbf{chebyshev heuristic} setting in $11 \times 11$ grid environment) utilizing each LLM-generated waypoint, as well as comparison with A* in number of explored states. The \textbf{blue} and \textbf{green rectangles} denote the start and goal states, respectively. \textbf{Grey rectangles} indicate the states explored by the \textbf{LLM-A*} algorithms, while \textbf{pink rectangles} represent states explored by A*. \textbf{Red line} illustrate the generated paths. \textbf{Stars} indicate LLM-generated waypoints. (See Section \ref{sec:qualitative} for more)
	\label{fig:showcase}}
\end{figure}

%% file: sections/5_conclusion.tex
\section{Conclusion}
In this work, we propose a novel path planning algorithm, \textbf{LLM-A*}, which outperforms traditional algorithms like A* in terms of both computational and memory efficiency, as well as LLM-only approach in path robustness and optimality. \textbf{LLM-A*} integrates heuristic values derived from LLM-generated waypoints (serves as global insight), with the deterministic guarantees in the A* algorithm. This hybrid approach addresses the shortcomings of both LLM-only approach and the A* algorithm by combining their respective strengths. Furthermore, the methodology of \textbf{LLM-A*} retains the general applicability of A*, making it suitable for pathfinding tasks in a wide range of environments. Thus, \textbf{LLM-A*} serves as an effective alternative to A* algorithm for path planning, especially in large-scale scenarios.

\section*{Limitations}
Although around 90\% of the paths generated by \textbf{LLM-A*} are optimal, our algorithm does not guarantee optimal path. While these cases are relatively few, they indicate that the algorithm may sometimes yield paths that are not the shortest or most efficient. Future improvements could focus on enhancing the optimality of the generated paths to ensure more consistent performance. 

Our experiments mainly utilized \textbf{GPT-3.5-TURBO} and \textbf{LLAMA3-8B-16bit} with basic prompt techniques. Although these models and prompts were adequate to validate the robustness of the \textbf{LLM-A*} algorithm, we did not explore a wider array of models or advanced prompt engineering strategies. Further testing with additional models and varied prompting methods could provide more comprehensive insights into the algorithm’s performance across different scenarios.

\section*{Acknowledgement}
The work is partially support by a DARPA ANSR program FA8750-23-2-0004 and a National Science Foundation CAREER award \#2339766. The views and conclusions are those of the authors and should not reflect the official policy or position of DARPA or the U.S. Government.

%% file: sections/6_appendix.tex
\section{Admissible Heuristic and Optimality}
In path planning algorithms such as A*, a heuristic function \( h(n) \) is deemed admissible if it never overestimates the cost to reach the goal from any given node \( n \). This ensures that the estimated cost from \( n \) to the goal does not exceed the actual lowest possible cost, thereby providing a lower bound on the true cost. An admissible heuristic guarantees that the A* algorithm will find an optimal solution, as it always explores the least costly path first.

The standard A* heuristic is often the Euclidean distance or straight-line distance between the current node and the goal, which is both admissible and consistent. This heuristic function accurately reflects the minimum possible cost in scenarios where there are no obstacles or other constraints that might alter the cost path.

However, the \textbf{LLM-A*} algorithm integrates an additional heuristic component, influenced by insights from large language models (LLMs), into the traditional A* heuristic function. Specifically, \textbf{LLM-A*} incorporates a modified heuristic \( h_{LLMA*}(n) \) which includes an additional cost term that estimates the difficulty of transitioning from the current state to the target state, based on the learned patterns from the LLM. This adjustment effectively amplifies the traditional heuristic by adding a factor derived from the LLM’s assessment of the state-space complexity and the likely transitions required.

Let \( h_{A*}(n) \) represent the conventional heuristic, and \( c_{LLM}(n) \) represent the cost component derived from the LLM insights. The modified heuristic can be expressed as:
\[
h_{LLMA*}(n) = h_{A*}(n) + c_{LLM}(n)
\]

The term \( c_{LLM}(n) \) may include factors such as predicted transition costs, obstacle avoidance strategies, or other environmental complexities inferred by the LLM, through selected target states in target list. Consequently, the heuristic function \( h_{LLMA*}(n) \) provides a more nuanced estimate of the cost to reach the goal, potentially guiding the search more effectively by leveraging the LLM's understanding of the domain.

While this enhanced heuristic expedites the search process by prioritizing paths that the LLM identifies as promising, it introduces a deviation from admissibility. By incorporating the additional cost \( c_{LLM}(n) \), the heuristic may overestimate the true cost to the goal, particularly if the LLM-derived costs are overly conservative or based on non-optimal path predictions. This overestimation violates the admissibility condition because the total estimated cost \( g(n) + h_{LLMA*}(n) \) could exceed the actual optimal path cost, where \( g(n) \) is the cost from the start to the current node.

The implications of this non-admissibility are significant: while the LLM-A* heuristic can potentially lead to faster convergence towards the goal by focusing the search in promising regions of the state space, it compromises the guarantee of finding the optimal path. The trade-off between search efficiency and optimality must be carefully considered in the application of LLM-A*. In scenarios where the heuristic insights from the LLM offer substantial benefits in reducing search time and computational resources, the potential loss of optimality may be justified. However, for applications where finding the absolute optimal path is crucial, relying solely on an admissible heuristic might be preferable.
\input{tables/prompt}
\section{Prompts in LLMs}
This appendix outlines the prompting techniques used in our LLM-A* algorithm to generate paths between start and goal points while navigating around obstacles. We employed different prompting strategies to evaluate their effectiveness in guiding the model. Below are the details of each technique along with the templates used.

\subsection{Standard 5-Shot Demonstration}

In the standard 5-shot demonstration in Table \ref{fig:fewshot}, the model is provided with five examples (or demonstrations) to guide the generation of the path. Each example includes start and goal points, along with horizontal and vertical barriers. The model is prompted to generate a path by following the pattern observed in the examples.

\subsection{Chain of Thought (CoT) Prompting}

The chain of thought prompting technique in Table \ref{fig:coT} provides a sequence of reasoning steps that the model follows to arrive at the final path. This technique includes a detailed thought process and evaluation for each step, helping the model to understand the rationale behind the path generation.

\subsection{Recursive Path Evaluation (RePE)}

In the recursive path evaluation technique shown Table \ref{fig:repe}, the model iteratively evaluates the path at each step and makes decisions based on previous iterations. This process involves selecting points, evaluating their effectiveness, and adjusting the path as necessary to avoid obstacles and reach the goal.

\section{Details of Dataset Construction}
The dataset for A* path planning is generated using a custom Python script, leveraging several key packages for randomization, geometric manipulation, visualization, and data management. The process involves the following steps:

\begin{enumerate}
    \item \textbf{Initialization}: The script initializes with specified map dimensions (x and y boundaries) and parameters (number of barriers and obstacles) for the number of unique environments and start-goal pairs.
    
    \item \textbf{Environment Creation}: For each map configuration, do the following:
    \begin{itemize}
        \item Random obstacles, horizontal barriers, and vertical barriers are generated within defined x and y ranges using the \texttt{shapely.geometry.LineString} for line segments.
        \item Start and goal points are randomly placed on the map, ensuring they do not intersect with any obstacles. Valid pairs form non-intersecting line segments.
    \end{itemize}
    
    \item \textbf{Data Storage}: The generated environments, including the obstacles and start-goal pairs, are stored in JSON format.
    
    \item \textbf{Query Generation}: Natural language queries are appended to each start-goal pair. These queries describe the task of finding a path that avoids the obstacles, which is supported as text input for LLMs.
    
    \item \textbf{Visualization}: The environments are visualized using \texttt{matplotlib}, displaying the grid, obstacles, and paths. The plots are supported to be saved as image files for reference and stream in a show..
\end{enumerate}

The Python packages utilized include:
\begin{itemize}
    \item \texttt{random}: For generating random coordinates.
    \item \texttt{shapely}: For geometric operations, specifically creating and validating the positions of obstacles and points.
    \item \texttt{matplotlib}: For plotting and saving visual representations of the environments.
    \item \texttt{inquirer}: For command-line prompts to make user decisions during dataset generation.
    \item \texttt{json} and \texttt{os}: For managing the reading and writing of dataset files.
    \item \texttt{search\_env}: A custom package for environment setup and plotting specific to the search based path planning task.
\end{itemize}
This process ensures a comprehensive dataset with varied environments and queries, suitable for training and testing A* path planning algorithms.

\section{Evaluation Metric}
In this study, we evaluate the performance of our algorithm using the geometric mean of ratios. This metric provides a robust measure for comparing the efficiency and effectiveness of different path planning algorithms. Below, we outline the rationale for choosing this metric, the calculation procedure, and its advantages.

\subsection{Rationale}

The geometric mean of ratios is used in this study to assess the relative performance of different path planning algorithms or approaches. It provides a balanced evaluation by aggregating multiple performance ratios, ensuring that no single extreme value disproportionately affects the overall metric. This is particularly useful in scenarios where the distribution of ratios can be skewed, and a simple arithmetic mean might be misleading.

\subsection{Calculation Procedure}

Let \( R_i \) represent the ratio of performance measures (such as path length, computation time, or any other relevant metric) between the proposed algorithm and a baseline or reference algorithm for the \( i \)-th test case. The geometric mean \( G \) of \( N \) ratios is calculated as follows:

\begin{equation}
G = \left( \prod_{i=1}^{N} R_i \right)^{\frac{1}{N}}
\end{equation}

The geometric mean \( G \) provides a multiplicative average, effectively normalizing the ratios and providing a single representative value that reflects the overall performance across all test cases.

\subsection{Advantages}

Using the geometric mean of ratios offers several benefits in the context of evaluating path planning algorithms:

\begin{enumerate}
    \item \textbf{Sensitivity to Relative Changes}: The geometric mean is sensitive to the relative differences between performance measures, making it suitable for comparing ratios.
    \item \textbf{Mitigation of Outliers}: Unlike the arithmetic mean, the geometric mean minimizes the impact of extreme values or outliers, providing a more stable and representative metric.
    \item \textbf{Interpretability}: The geometric mean allows for easy interpretation of performance improvements or deteriorations. A geometric mean greater than 1 indicates that, on average, the proposed algorithm performs better than the baseline, while a value less than 1 suggests poorer performance.
    \item \textbf{Scalability}: The geometric mean naturally scales with multiplicative factors, making it appropriate for comparing algorithms across different scales or units of measurement.
\end{enumerate}

%% file: tables/prompt.tex
\begin{table*}[tb!]
    \centering
    \fontsize{10pt}{12pt}\selectfont
    \fbox{
        \begin{minipage}{0.95\textwidth}
        Identify a path between the start and goal points to navigate around obstacles and find the shortest path to the goal. 
Horizontal barriers are represented as [y, x\_start, x\_end], and vertical barriers are represented as [x, y\_start, y\_end].
Conclude your response with the generated path in the format "Generated Path: [[x1, y1], [x2, y2], ...]".
        \vspace{0.3cm} 

        Start Point: [5, 5]\\
        Goal Point: [20, 20]\\
        Horizontal Barriers: [[10, 0, 25], [15, 30, 50]]\\
        Vertical Barriers: [[25, 10, 22]]\\
        Generated Path: [[5, 5], [26, 9], [25, 23], [20, 20]]

        \vspace{0.3cm}
        \textcolor{gray}{[5 in-context demonstrations abbreviated]}
        \vspace{0.3cm}

        Start Point: \{start\}\\
        Goal Point: \{goal\}\\
        Horizontal Barriers: \{horizontal\_barriers\}\\
        Vertical Barriers: \{vertical\_barriers\}\\
        Generated Path: \underline{\textbf{Model Generated Answer Goes Here}}
        \end{minipage}
    }
    \caption{The template of the prompt we used for LLM-A* using standard 5-shot demonstration.}
    \label{fig:fewshot}
\end{table*}

\begin{table*}[tb!]
    \centering
    \fontsize{10pt}{12pt}\selectfont
    \fbox{
        \begin{minipage}{0.95\textwidth}
Identify a path between the start and goal points to navigate around obstacles and find the shortest path to the goal. 
Horizontal barriers are represented as [y, x\_start, x\_end], and vertical barriers are represented as [x, y\_start, y\_end].
Conclude your response with the generated path in the format "Generated Path: [[x1, y1], [x2, y2], ...]".
        \vspace{0.3cm} 

Start Point: [5, 5] \\
Goal Point: [20, 20] \\
Horizontal Barriers: [[10, 0, 25], [15, 30, 50]] \\
Vertical Barriers: [[25, 10, 22]] \\
Thought: Identify a path from [5, 5] to [20, 20] while avoiding the horizontal barrier at y=10 spanning x=0 to x=25 by moving upwards and right, then bypass the vertical barrier at x=25 spanning y=10 to y=22, and finally move directly to [20, 20]. \\
Generated Path: [[5, 5], [26, 9], [25, 23], [20, 20]]

        \vspace{0.3cm}
        \textcolor{gray}{[3 in-context demonstrations abbreviated]}
        \vspace{0.3cm}

        Start Point: \{start\}\\
        Goal Point: \{goal\}\\
        Horizontal Barriers: \{horizontal\_barriers\}\\
        Vertical Barriers: \{vertical\_barriers\}\\
        Generated Path: \underline{\textbf{Model Generated Answer Goes Here}}
        \end{minipage}
    }
    \caption{The template of the prompt we used for LLM-A* using standard 3-shot demonstration with chain of thought generation process.}
    \label{fig:coT}
\end{table*}

\begin{table*}[tb!]
    \centering
    \fontsize{10pt}{12pt}\selectfont
    \fbox{
        \begin{minipage}{0.95\textwidth}
Identify a path between the start and goal points to navigate around obstacles and find the shortest path to the goal. 
Horizontal barriers are represented as [y, x\_start, x\_end], and vertical barriers are represented as [x, y\_start, y\_end].
Conclude your response with the generated path in the format "Generated Path: [[x1, y1], [x2, y2], ...]".
        \vspace{0.3cm} 

Start Point: [5, 5] \\
Goal Point: [20, 20] \\
Horizontal Barriers: [[10, 0, 25], [15, 30, 50]] \\
Vertical Barriers: [[25, 10, 22]] \\
- First Iteration on [5, 5] \\
Thought: The horizontal barrier at y=10 spanning x=0 to x=25 blocks the direct path to the goal. To navigate around it, we should move to the upper-right corner of the barrier. \\
Selected Point: [26, 9] \\
Evaluation: The selected point [26, 9] effectively bypasses the horizontal barrier, positioning us at its corner and maintaining progress toward the goal without encountering additional obstacles. \\
- Second Iteration on [26, 9] \\
Thought: Now that we have bypassed the horizontal barrier, the path to the goal seems clear. \\
Selected Point: [20, 20] \\
Evaluation: The path is obstructed by the vertical barrier, leading to a collision. A more effective route involves moving around this vertical barrier. \\
Thought: To bypass the vertical barrier at x=25, we should move along its length and then turn around it to continue toward the goal. \\
Selected Point: [25, 23] \\
Evaluation: The selected point [25, 23] successfully avoids the vertical barrier and brings us closer to the goal without encountering further obstacles. \\
- Third Iteration on [25, 23] \\
Thought: From this position, there are no barriers directly obstructing the path to the goal. \\
Selected Point: [20, 20] \\
Evaluation: The path to the goal is clear from here, allowing a direct move to the goal. \\
Generated Path: [[5, 5], [26, 9], [25, 23], [20, 20]]

        \vspace{0.3cm}
        \textcolor{gray}{[3 in-context demonstrations abbreviated]}
        \vspace{0.3cm}

        Start Point: \{start\}\\
        Goal Point: \{goal\}\\
        Horizontal Barriers: \{horizontal\_barriers\}\\
        Vertical Barriers: \{vertical\_barriers\}\\
        Generated Path: \underline{\textbf{Model Generated Answer Goes Here}}
        \end{minipage}
    }
    \caption{The template of the prompt we used for LLM-A* using standard 3-shot demonstration with recursive path evaluation generation process.}
    \label{fig:repe}
\end{table*}

%% file: acl2024.bbl
\begin{thebibliography}{56}
\expandafter\ifx\csname natexlab\endcsname\relax\def\natexlab#1{#1}\fi

\bibitem[{Abdou et~al.(2021)Abdou, Kulmizev, Hershcovich, Frank, Pavlick, and S{\o}gaard}]{abdou2021can}
Mostafa Abdou, Artur Kulmizev, Daniel Hershcovich, Stella Frank, Ellie Pavlick, and Anders S{\o}gaard. 2021.
\newblock Can language models encode perceptual structure without grounding? a case study in color.
\newblock \emph{arXiv preprint arXiv:2109.06129}.

\bibitem[{Aghzal et~al.(2023)Aghzal, Plaku, and Yao}]{aghzal2023can}
Mohamed Aghzal, Erion Plaku, and Ziyu Yao. 2023.
\newblock Can large language models be good path planners? a benchmark and investigation on spatial-temporal reasoning.
\newblock \emph{arXiv preprint arXiv:2310.03249}.

\bibitem[{Agrawal(2023)}]{agrawal2023llms}
Shrivats Agrawal. 2023.
\newblock Are llms the master of all trades?: Exploring domain-agnostic reasoning skills of llms.
\newblock \emph{arXiv preprint arXiv:2303.12810}.

\bibitem[{Botea et~al.(2004)Botea, Müller, and Schaeffer}]{botea2004near}
Adi Botea, Martin Müller, and Jonathan Schaeffer. 2004.
\newblock Near optimal hierarchical path-finding.
\newblock \emph{Journal of Game Development}, 1(1):7--28.

\bibitem[{Bubeck et~al.(2023)Bubeck, Chandrasekaran, Eldan, Gehrke, Horvitz, Kamar, Lee, Lee, Li, Lundberg et~al.}]{bubeck2023sparks}
S{\'e}bastien Bubeck, Varun Chandrasekaran, Ronen Eldan, Johannes Gehrke, Eric Horvitz, Ece Kamar, Peter Lee, Yin~Tat Lee, Yuanzhi Li, Scott Lundberg, et~al. 2023.
\newblock Sparks of artificial general intelligence: Early experiments with gpt-4.
\newblock \emph{arXiv preprint arXiv:2303.12712}.

\bibitem[{Cao et~al.(2019)Cao, Law, and Fidler}]{cao2019theoretical}
Tianshi Cao, Marc Law, and Sanja Fidler. 2019.
\newblock A theoretical analysis of the number of shots in few-shot learning.
\newblock \emph{arXiv preprint arXiv:1909.11722}.

\bibitem[{Chen et~al.(2023{\natexlab{a}})Chen, Shu, Shareghi, Collier, Narasimhan, and Yao}]{chen2023fireact}
Baian Chen, Chang Shu, Ehsan Shareghi, Nigel Collier, Karthik Narasimhan, and Shunyu Yao. 2023{\natexlab{a}}.
\newblock Fireact: Toward language agent fine-tuning.
\newblock \emph{arXiv preprint arXiv:2310.05915}.

\bibitem[{Chen et~al.(2023{\natexlab{b}})Chen, Arkin, Zhang, Roy, and Fan}]{chen2023autotamp}
Yongchao Chen, Jacob Arkin, Yang Zhang, Nicholas Roy, and Chuchu Fan. 2023{\natexlab{b}}.
\newblock Autotamp: Autoregressive task and motion planning with llms as translators and checkers.
\newblock \emph{arXiv preprint arXiv:2306.06531}.

\bibitem[{C{\^o}t{\'e} et~al.(2019)C{\^o}t{\'e}, K{\'a}d{\'a}r, Yuan, Kybartas, Barnes, Fine, Moore, Hausknecht, El~Asri, Adada et~al.}]{cote2019textworld}
Marc-Alexandre C{\^o}t{\'e}, Akos K{\'a}d{\'a}r, Xingdi Yuan, Ben Kybartas, Tavian Barnes, Emery Fine, James Moore, Matthew Hausknecht, Layla El~Asri, Mahmoud Adada, et~al. 2019.
\newblock Textworld: A learning environment for text-based games.
\newblock In \emph{Computer Games: 7th Workshop, CGW 2018, Held in Conjunction with the 27th International Conference on Artificial Intelligence, IJCAI 2018, Stockholm, Sweden, July 13, 2018, Revised Selected Papers 7}, pages 41--75. Springer.

\bibitem[{Demyen and Buro(2006)}]{demyen2006efficient}
Douglas Demyen and Michael Buro. 2006.
\newblock Efficient triangulation-based pathfinding.
\newblock In \emph{Proceedings of the AAAI Conference on Artificial Intelligence}, pages 942--947.

\bibitem[{Ding et~al.(2024)Ding, Fang, Li, Wang, Zhou, Yu, Li, Walter, and Mei}]{ding2024mango}
Peng Ding, Jiading Fang, Peng Li, Kangrui Wang, Xiaochen Zhou, Mo~Yu, Jing Li, Matthew~R Walter, and Hongyuan Mei. 2024.
\newblock Mango: A benchmark for evaluating mapping and navigation abilities of large language models.
\newblock \emph{arXiv preprint arXiv:2403.19913}.

\bibitem[{Dou et~al.(2024)Dou, Yang, Wu, Chang, and Peng}]{dou2024reflection}
Zi-Yi Dou, Cheng-Fu Yang, Xueqing Wu, Kai-Wei Chang, and Nanyun Peng. 2024.
\newblock Reflection-reinforced self-training for language agents.
\newblock \emph{arXiv preprint arXiv:2406.01495}.

\bibitem[{Fiorini and Shiller(1998)}]{fiorini1998motion}
Paolo Fiorini and Zvi Shiller. 1998.
\newblock Motion planning in dynamic environments using velocity obstacles.
\newblock In \emph{IEEE International Conference on Robotics and Automation}, pages 760--765. IEEE.

\bibitem[{Fox et~al.(1997)Fox, Burgard, and Thrun}]{fox1997dynamic}
Dieter Fox, Wolfram Burgard, and Sebastian Thrun. 1997.
\newblock The dynamic window approach to collision avoidance.
\newblock \emph{IEEE Robotics \& Automation Magazine}, 4(1):23--33.

\bibitem[{Harabor and Grastien(2011)}]{harabor2011online}
Daniel Harabor and Alban Grastien. 2011.
\newblock Online graph pruning for pathfinding on grid maps.
\newblock In \emph{Proceedings of the AAAI conference on artificial intelligence}, volume~25, pages 1114--1119.

\bibitem[{Hart et~al.(1968{\natexlab{a}})Hart, Nilsson, and Raphael}]{hart1968a}
Peter Hart, Nils Nilsson, and Bertram Raphael. 1968{\natexlab{a}}.
\newblock A formal basis for the heuristic determination of minimum cost paths.
\newblock \emph{IEEE Transactions on Systems Science and Cybernetics}, 4(2):100--107.

\bibitem[{Hart et~al.(1968{\natexlab{b}})Hart, Nilsson, and Raphael}]{hart1968formal}
Peter~E Hart, Nils~J Nilsson, and Bertram Raphael. 1968{\natexlab{b}}.
\newblock A formal basis for the heuristic determination of minimum cost paths.
\newblock \emph{IEEE transactions on Systems Science and Cybernetics}, 4(2):100--107.

\bibitem[{Holte et~al.(1996)Holte, Perez, Zimmer, and MacDonald}]{holte1996hierarchical}
Robert Holte, M~Perez, R~Zimmer, and A~MacDonald. 1996.
\newblock Hierarchical a\*.
\newblock In \emph{Proceedings of the AAAI Conference on Artificial Intelligence}, pages 530--535.

\bibitem[{Ilharco et~al.(2020)Ilharco, Zellers, Farhadi, and Hajishirzi}]{ilharco2020probing}
Gabriel Ilharco, Rowan Zellers, Ali Farhadi, and Hannaneh Hajishirzi. 2020.
\newblock Probing contextual language models for common ground with visual representations.
\newblock \emph{arXiv preprint arXiv:2005.00619}.

\bibitem[{Jansen and Buro(2007)}]{jansen2007hpa}
M~Jansen and Michael Buro. 2007.
\newblock Hpa* enhancements.
\newblock In \emph{Proceedings of the AAAI Conference on Artificial Intelligence and Interactive Digital Entertainment}, volume~3, pages 84--87.

\bibitem[{Karaman and Frazzoli(2011)}]{karaman2011sampling}
Sertac Karaman and Emilio Frazzoli. 2011.
\newblock Sampling-based algorithms for optimal motion planning.
\newblock \emph{The International Journal of Robotics Research}, 30(7):846--894.

\bibitem[{Koch(2011)}]{koch2011grid}
Uwe Koch. 2011.
\newblock Grid-specific feature of hpa*.
\newblock In \emph{Proceedings of the International Conference on Artificial Intelligence}, pages 135--142.

\bibitem[{Koenig et~al.(2004)Koenig, Likhachev, and Furcy}]{koenig2004lifelong}
Sven Koenig, Maxim Likhachev, and David Furcy. 2004.
\newblock Lifelong planning a\*.
\newblock \emph{Artificial Intelligence}, 155(1-2):93--146.

\bibitem[{Korf(1985)}]{korf1985depth}
Richard~E Korf. 1985.
\newblock Depth-first iterative-deepening: An optimal admissible tree search.
\newblock \emph{Artificial Intelligence}, 27(1):97--109.

\bibitem[{Korf(1990)}]{korf1990real}
Richard~E Korf. 1990.
\newblock Real-time heuristic search.
\newblock \emph{Artificial Intelligence}, 42(2-3):189--211.

\bibitem[{Korf et~al.(2001)Korf, Reid, and Edelkamp}]{korf2001time}
Richard~E Korf, Michael Reid, and Stefan Edelkamp. 2001.
\newblock Time complexity of iterative-deepening-a*.
\newblock \emph{Artificial Intelligence}, 129(1-2):199--218.

\bibitem[{Latif(2024)}]{latif20243p}
Ehsan Latif. 2024.
\newblock 3p-llm: Probabilistic path planning using large language model for autonomous robot navigation.
\newblock \emph{arXiv preprint arXiv:2403.18778}.

\bibitem[{LaValle(2006)}]{lavalle2006planning}
Steven~M LaValle. 2006.
\newblock \emph{Planning Algorithms}.
\newblock Cambridge University Press.

\bibitem[{Liu et~al.(2023)Liu, Jiang, Zhang, Liu, Zhang, Biswas, and Stone}]{liu2023llm+}
Bo~Liu, Yuqian Jiang, Xiaohan Zhang, Qiang Liu, Shiqi Zhang, Joydeep Biswas, and Peter Stone. 2023.
\newblock Llm+ p: Empowering large language models with optimal planning proficiency.
\newblock \emph{arXiv preprint arXiv:2304.11477}.

\bibitem[{Nash et~al.(2007)Nash, Daniel, Koenig, and Felner}]{nash2007theta}
Alex Nash, Kenny Daniel, Sven Koenig, and Ariel Felner. 2007.
\newblock Theta\*: Any-angle path planning on grids.
\newblock In \emph{Proceedings of the AAAI Conference on Artificial Intelligence}, pages 1177--1183.

\bibitem[{Naveed et~al.(2023)Naveed, Khan, Qiu, Saqib, Anwar, Usman, Barnes, and Mian}]{naveed2023comprehensive}
Humza Naveed, Asad~Ullah Khan, Shi Qiu, Muhammad Saqib, Saeed Anwar, Muhammad Usman, Nick Barnes, and Ajmal Mian. 2023.
\newblock A comprehensive overview of large language models.
\newblock \emph{arXiv preprint arXiv:2307.06435}.

\bibitem[{Patel and Pavlick(2021)}]{patel2021mapping}
Roma Patel and Ellie Pavlick. 2021.
\newblock Mapping language models to grounded conceptual spaces.
\newblock In \emph{International Conference on Learning Representations}.

\bibitem[{Pearl(1984)}]{pearl1984heuristics}
Judea Pearl. 1984.
\newblock \emph{Heuristics: Intelligent Search Strategies for Computer Problem Solving}.
\newblock Addison-Wesley.

\bibitem[{Razeghi et~al.(2022)Razeghi, Logan~IV, Gardner, and Singh}]{razeghi2022impact}
Yasaman Razeghi, Robert~L Logan~IV, Matt Gardner, and Sameer Singh. 2022.
\newblock Impact of pretraining term frequencies on few-shot numerical reasoning.
\newblock In \emph{Findings of the Association for Computational Linguistics: EMNLP 2022}, pages 840--854.

\bibitem[{Renze and Guven(2024)}]{renze2024self}
Matthew Renze and Erhan Guven. 2024.
\newblock Self-reflection in llm agents: Effects on problem-solving performance.
\newblock \emph{arXiv preprint arXiv:2405.06682}.

\bibitem[{Ruis et~al.(2020)Ruis, Andreas, Baroni, Bouchacourt, and Lake}]{ruis2020benchmark}
Laura Ruis, Jacob Andreas, Marco Baroni, Diane Bouchacourt, and Brenden~M Lake. 2020.
\newblock A benchmark for systematic generalization in grounded language understanding.
\newblock \emph{Advances in neural information processing systems}, 33:19861--19872.

\bibitem[{Russell(1992)}]{russell1992memory}
Stuart~J Russell. 1992.
\newblock Memory-bounded heuristic search.
\newblock \emph{Artificial Intelligence}, 49(1-3):5--27.

\bibitem[{Shah et~al.(2023)Shah, Equi, Osi{\'n}ski, Xia, Ichter, and Levine}]{shah2023navigation}
Dhruv Shah, Michael~Robert Equi, B{\l}a{\.z}ej Osi{\'n}ski, Fei Xia, Brian Ichter, and Sergey Levine. 2023.
\newblock Navigation with large language models: Semantic guesswork as a heuristic for planning.
\newblock In \emph{Conference on Robot Learning}, pages 2683--2699. PMLR.

\bibitem[{Shinn et~al.(2024)Shinn, Cassano, Gopinath, Narasimhan, and Yao}]{shinn2024reflexion}
Noah Shinn, Federico Cassano, Ashwin Gopinath, Karthik Narasimhan, and Shunyu Yao. 2024.
\newblock Reflexion: Language agents with verbal reinforcement learning.
\newblock \emph{Advances in Neural Information Processing Systems}, 36.

\bibitem[{Shridhar et~al.(2020{\natexlab{a}})Shridhar, Thomason, Gordon, Bisk, Han, Mottaghi, Zettlemoyer, and Fox}]{shridhar2020alfred}
Mohit Shridhar, Jesse Thomason, Daniel Gordon, Yonatan Bisk, Winson Han, Roozbeh Mottaghi, Luke Zettlemoyer, and Dieter Fox. 2020{\natexlab{a}}.
\newblock Alfred: A benchmark for interpreting grounded instructions for everyday tasks.
\newblock In \emph{Proceedings of the IEEE/CVF conference on computer vision and pattern recognition}, pages 10740--10749.

\bibitem[{Shridhar et~al.(2020{\natexlab{b}})Shridhar, Yuan, C{\^o}t{\'e}, Bisk, Trischler, and Hausknecht}]{shridhar2020alfworld}
Mohit Shridhar, Xingdi Yuan, Marc-Alexandre C{\^o}t{\'e}, Yonatan Bisk, Adam Trischler, and Matthew Hausknecht. 2020{\natexlab{b}}.
\newblock Alfworld: Aligning text and embodied environments for interactive learning.
\newblock \emph{arXiv preprint arXiv:2010.03768}.

\bibitem[{Silver et~al.(2022)Silver, Hariprasad, Shuttleworth, Kumar, Lozano-P{\'e}rez, and Kaelbling}]{silver2022pddl}
Tom Silver, Varun Hariprasad, Reece~S Shuttleworth, Nishanth Kumar, Tom{\'a}s Lozano-P{\'e}rez, and Leslie~Pack Kaelbling. 2022.
\newblock Pddl planning with pretrained large language models.
\newblock In \emph{NeurIPS 2022 foundation models for decision making workshop}.

\bibitem[{Song et~al.(2023)Song, Wu, Washington, Sadler, Chao, and Su}]{song2023llm}
Chan~Hee Song, Jiaman Wu, Clayton Washington, Brian~M Sadler, Wei-Lun Chao, and Yu~Su. 2023.
\newblock Llm-planner: Few-shot grounded planning for embodied agents with large language models.
\newblock In \emph{Proceedings of the IEEE/CVF International Conference on Computer Vision}, pages 2998--3009.

\bibitem[{Stentz(1994)}]{stentz1994optimal}
Anthony Stentz. 1994.
\newblock Optimal and efficient path planning for partially-known environments.
\newblock In \emph{Proceedings of the IEEE International Conference on Robotics and Automation (ICRA)}, pages 3310--3317.

\bibitem[{Thrun et~al.(2005)Thrun, Burgard, and Fox}]{thrun2005probabilistic}
Sebastian Thrun, Wolfram Burgard, and Dieter Fox. 2005.
\newblock \emph{Probabilistic Robotics}.
\newblock MIT press.

\bibitem[{Wang et~al.(2024)Wang, Han, Jiao, Zhang, Wu, Zhu, and Liu}]{wang2024llm}
Shu Wang, Muzhi Han, Ziyuan Jiao, Zeyu Zhang, Ying~Nian Wu, Song-Chun Zhu, and Hangxin Liu. 2024.
\newblock Llm\^{} 3: Large language model-based task and motion planning with motion failure reasoning.
\newblock \emph{arXiv preprint arXiv:2403.11552}.

\bibitem[{Wei et~al.(2022)Wei, Wang, Schuurmans, Bosma, Xia, Chi, Le, Zhou et~al.}]{wei2022chain}
Jason Wei, Xuezhi Wang, Dale Schuurmans, Maarten Bosma, Fei Xia, Ed~Chi, Quoc~V Le, Denny Zhou, et~al. 2022.
\newblock Chain-of-thought prompting elicits reasoning in large language models.
\newblock \emph{Advances in Neural Information Processing Systems}, 35:24824--24837.

\bibitem[{Wu et~al.(2023)Wu, Qiu, Ross, Aky{\"u}rek, Chen, Wang, Kim, Andreas, and Kim}]{wu2023reasoning}
Zhaofeng Wu, Linlu Qiu, Alexis Ross, Ekin Aky{\"u}rek, Boyuan Chen, Bailin Wang, Najoung Kim, Jacob Andreas, and Yoon Kim. 2023.
\newblock Reasoning or reciting? exploring the capabilities and limitations of language models through counterfactual tasks.
\newblock \emph{arXiv preprint arXiv:2307.02477}.

\bibitem[{Wu et~al.(2021)Wu, Kreiss, Ong, and Potts}]{wu2021reascan}
Zhengxuan Wu, Elisa Kreiss, Desmond~C Ong, and Christopher Potts. 2021.
\newblock Reascan: Compositional reasoning in language grounding.
\newblock \emph{arXiv preprint arXiv:2109.08994}.

\bibitem[{Xie et~al.(2023)Xie, Yu, Zhu, Bai, Gong, and Soh}]{xie2023translating}
Yaqi Xie, Chen Yu, Tongyao Zhu, Jinbin Bai, Ze~Gong, and Harold Soh. 2023.
\newblock Translating natural language to planning goals with large-language models.
\newblock \emph{arXiv preprint arXiv:2302.05128}.

\bibitem[{Yang et~al.(2023{\natexlab{a}})Yang, Chen, Yang, Dai, Yuan, Wang, and Chang}]{yang2023lacma}
Cheng-Fu Yang, Yen-Chun Chen, Jianwei Yang, Xiyang Dai, Lu~Yuan, Yu-Chiang~Frank Wang, and Kai-Wei Chang. 2023{\natexlab{a}}.
\newblock Lacma: Language-aligning contrastive learning with meta-actions for embodied instruction following.
\newblock \emph{arXiv preprint arXiv:2310.12344}.

\bibitem[{Yang et~al.(2023{\natexlab{b}})Yang, Xu, Wu, Gao, Chang, and Gao}]{yang2023planning}
Cheng-Fu Yang, Haoyang Xu, Te-Lin Wu, Xiaofeng Gao, Kai-Wei Chang, and Feng Gao. 2023{\natexlab{b}}.
\newblock Planning as in-painting: A diffusion-based embodied task planning framework for environments under uncertainty.
\newblock \emph{arXiv preprint arXiv:2312.01097}.

\bibitem[{Yao et~al.(2022)Yao, Zhao, Yu, Du, Shafran, Narasimhan, and Cao}]{yao2022react}
Shunyu Yao, Jeffrey Zhao, Dian Yu, Nan Du, Izhak Shafran, Karthik Narasimhan, and Yuan Cao. 2022.
\newblock React: Synergizing reasoning and acting in language models.
\newblock \emph{arXiv preprint arXiv:2210.03629}.

\bibitem[{Yin et~al.(2023)Yin, Brahman, Ravichander, Chandu, Chang, Choi, and Lin}]{yin2023lumos}
Da~Yin, Faeze Brahman, Abhilasha Ravichander, Khyathi Chandu, Kai-Wei Chang, Yejin Choi, and Bill~Yuchen Lin. 2023.
\newblock Lumos: Learning agents with unified data, modular design, and open-source llms.
\newblock \emph{arXiv preprint arXiv:2311.05657}.

\bibitem[{Zheng et~al.(2023)Zheng, Mishra, Chen, Cheng, Chi, Le, and Zhou}]{zheng2023take}
Huaixiu~Steven Zheng, Swaroop Mishra, Xinyun Chen, Heng-Tze Cheng, Ed~H Chi, Quoc~V Le, and Denny Zhou. 2023.
\newblock Take a step back: Evoking reasoning via abstraction in large language models.
\newblock \emph{arXiv preprint arXiv:2310.06117}.

\bibitem[{Zhou et~al.(2024)Zhou, Hong, and Wu}]{zhou2024navgpt}
Gengze Zhou, Yicong Hong, and Qi~Wu. 2024.
\newblock Navgpt: Explicit reasoning in vision-and-language navigation with large language models.
\newblock In \emph{Proceedings of the AAAI Conference on Artificial Intelligence}, volume~38, pages 7641--7649.

\end{thebibliography}
